\documentclass[lettersize,journal]{IEEEtran}
\usepackage{amsmath,amsfonts}
\usepackage{algorithmic}
\usepackage{algorithm}
\usepackage{array}
\usepackage[caption=false,font=normalsize,labelfont=sf,textfont=sf]{subfig}
\usepackage[T1]{fontenc}
\usepackage{textcomp}
\usepackage{stfloats}
\usepackage{url}
\usepackage{verbatim}
\usepackage{graphicx}
\usepackage{cite}
\usepackage{booktabs}
\usepackage{pifont}
\usepackage{xcolor}
\usepackage{balance}

\begin{document}

\title{Next-Frame Feature Prediction for Multimodal Deepfake Detection and Temporal Localization}

\author{
    Ashutosh Anshul, Shreyas Gopal, Deepu Rajan, and Eng Siong Chng\\
    \textit{College of Computing and Data Science} \\
    \textit{Nanyang Technological University, Singapore}\\
}



\maketitle

\begin{abstract}
Recent multimodal deepfake detection methods designed for generalization conjecture that single-stage supervised training struggles to generalize across unseen manipulations and datasets. However, such approaches that target generalization require pretraining over real samples. Additionally, these methods primarily focus on detecting audio-visual inconsistencies and may overlook intra-modal artifacts causing them to fail against manipulations that preserve audio-visual alignment. To address these limitations, we propose a single-stage training framework that enhances generalization by incorporating next-frame prediction for both uni-modal and cross-modal features. Additionally, we introduce a window-level attention mechanism to capture discrepancies between predicted and actual frames, enabling the model to detect local artifacts around every frame, which is crucial for accurately classifying fully manipulated videos and effectively localizing deepfake segments in partially spoofed samples. Our model, evaluated on multiple benchmark datasets, demonstrates strong generalization and precise temporal localization.
\end{abstract}

\begin{IEEEkeywords}
Deepfake Detection, Multimodal, Generalization, Temporal Localization, Next frame Prediction
\end{IEEEkeywords}    
\section{Introduction}
\label{sec:intro}

Rapid advancements in generative modeling now enable the creation of increasingly realistic fake media and of new kinds of manipulations, posing risks of fraud, identity theft, and misinformation. Hence, developing robust detection models that generalize to unseen manipulations and data distributions is crucial as generative AI continues to evolve. Moreover, most existing audio-visual deepfake detection models assume manipulation occurs throughout an entire video, while in reality, it can appear in small, noncontiguous segments. In such cases, precise temporal localization is needed instead of a simple binary classification of the video.

This paper proposes a multimodal framework for deepfake detection and temporal localization. Existing audio-visual methods typically fuse audio and visual features~\cite{astrid2024detecting, kharel2023df, wang2024avt2}, with some incorporating contrastive loss to strengthen cross-modal correspondence~\cite{yang2023avoid, zou2024cross, liu2023mcl}. Most approaches rely on single-stage supervised training~\cite{zhou2021joint}, which, as~\cite{oorloff2024avff} argues, restricts the ability to fully capture audio-visual consistencies and reduces generalization to unseen manipulations. Recent work shows that pretraining on real samples with multimodal learning can improve detection by capturing subtle cross-modal inconsistencies~\cite{feng2023self, haliassos2022leveraging, oorloff2024avff}. However, these methods face one or more of the following limitations: (i) the need for two-stage training to achieve robust generalization, (ii) the emphasis only on cross-modal features while potentially overlooking intra-modal inconsistencies, making them vulnerable to attacks that preserve audio-visual alignment but introduce inconsistencies within a single modality, and (iii) limited adaptability for temporal localization. MRDF~\cite{zou2024cross} attempts to address both modality-specific and cross-modal cues in a single-stage framework by introducing unimodal regularization losses during training. However, the unimodal heads are discarded at inference, and only the cross-modal transformer is used, assuming that modality-specific inconsistencies are retained in the fused representation. This reliance may prevent the model from explicitly leveraging unimodal signals.

To address these challenges, we propose a single-stage training framework that generalizes to unseen manipulations and adapts to both classification and temporal localization. Recent studies~\cite{zhang2023ummaformer, zhang2021unsupervised, zhang2019deep} show that feature reconstruction is effective for anomaly detection in time-series data. UMMAFormer~\cite{zhang2023ummaformer} utilizes an encoder-decoder based Temporal Feature Abnormal Attention (TFAA) to capture abnormalities in temporal features and shows its positive contribution on its performance. Motivated by the results and their impact on UMMAFormer~\cite{zhang2023ummaformer}, we conjecture that the difference between reconstructed and actual features can be helpful in generalizable deepfake detection. Building on this, we introduce the Masked-Prediction Feature Extraction Module, a core component of our approach. This module predicts the next frame’s features from previous frames and explicitly captures discrepancies between predicted and actual features, enabling the detection of local inconsistencies. Since manipulations often appear in short segments rather than across the full video, focusing on local information around each frame is critical for accurate localization while preserving classification performance. Hence, we apply local window-based attention, which allows the model to compare predicted and actual features in a localized manner. Unlike prior work that relies only on cross-modal features~\cite{oorloff2024avff, feng2023self}, or MRDF~\cite{zou2024cross}, which discards unimodal cues at inference, we employ three masked-prediction-based modules, two unimodal and one cross-modal, to capture both intra- and inter-modal inconsistencies explicitly. Classification and localization tasks are handled by separate prediction heads but share a common backbone, making the framework adaptable to both. Finally, inspired by contrastive learning, we apply a frame-level contrastive loss between predicted and actual frame features to guide the model in learning discriminative unimodal and cross-modal representations.

We evaluate our approach against existing methods across multiple datasets and settings. Despite being a single-stage framework, the proposed method generalizes well to unseen manipulations and different deepfake datasets. Unlike other single-stage multimodal detection methods, it maintains consistent performance across all cross-manipulation categories. By leveraging differences between predicted and actual frame-level features, our model captures both intra-modal and cross-modal inconsistencies for deepfake classification and temporal localization. Analysis of the gap between predicted and actual features shows that the Masked Prediction approach also enhances model interpretability and can help explain both detection and temporal localization decisions. In summary:

\begin{itemize} 
    \item We propose a Masked-Prediction Feature Extraction Module that predicts the next frame's features using previous frames and measures discrepancies between predicted and actual frames. This helps detect local inconsistencies around each frame, making the model effective for both deepfake classification and temporal localization. 
    \item To capture fine-grained inconsistencies, we use local window-based attention to compare predicted and actual frame features. This enables the model to focus on local variations, improving its ability to detect subtle manipulations without affecting classification performance. 
    \item Unlike previous methods that rely solely on cross-modal inconsistencies, we propose to use three masked-prediction modules. This joint learning approach ensures the model captures both within-modality and cross-modality discrepancies.
    \item Our design shows strong generalization across unseen manipulations and datasets while maintaining a single-stage framework, achieving performance comparable to state-of-the-art methods for both classification and localization. Our model is adaptable for both classification and temporal localization using a shared backbone.
\end{itemize}
\section{Related Works}
\label{sec:related_works}

\subsection{Deepfake Detection}
\textbf{Visual-only Approaches:} Most of the visual-only proposals focused on detecting manipulations around the face region. While some methods focus on spatiotemporal~\cite{pang2023mre, wang2023altfreezing, zhang2024learning, yu2023augmented} and temporal~\cite{choi2024exploiting, zheng2021exploring, ge2022deepfake} inconsistencies, others look for defects in face blending~\cite{li2020face}, and texture~\cite{liu2020global, zhao2021multi}. Some of the approaches try to spot irregularities from the frequency domain~\cite{10286083, gu2022exploiting, jeong2022frepgan, liu2024hierarchical, 10107603}. Some visual approaches detect inconsistencies in noise patterns~\cite{chen2024compressed, wang2023noise, qiao2024deepfake} or semantics~\cite{xu2023tall, xu2024towards}, while others proposed identity based approaches for deepfake detection~\cite{cozzolino2021id, huang2023implicit}.  XceptionNet, a convolution-based network, is proposed by~\cite{rossler2019faceforensics++} as a benchmark for their FaceForensics++ dataset. However, it does not explore the generalization capability of the model. Recently, researchers have been focusing on generalizability in video deepfake detection~\cite{yan2024transcending, tan2024rethinking, 11045799, lin2024preserving, 10516609, nie2024dip, 11098842, 10654318}. LipForensics~\cite{haliassos2021lips} introduces a generalizable deepfake detection method that identifies irregularities in lip movement. FTCN ~\cite{zheng2021exploring}, proposed by Zheng et al., is a two-stage end-to-end network combining a convolutional network with a transformer to analyze facial temporal coherence. RealForensics~\cite{haliassos2022leveraging} employs audio-visual pretraining, but its classification stage discards audio and relies solely on visual deepfake detection.~\cite{wang2022deepfake} improves generalization by training on adversarially crafted samples designed to attack classification models through methods that blur high-frequency artifacts in facial manipulations.~\cite{prashnani2025generalizable} enhances generalization by leveraging temporal variations in phase information within the frequency domain of facial regions. Although the visual-only approaches are quite effective against facial manipulations, their applicability and effectiveness against deepfakes that involve multimodal manipulations are limited.

\textbf{Audio-Visual Approaches:} End-to-end supervised training on deepfake labels has been explored in various studies. For example,~\cite{chugh2020not} employs contrastive loss to model dissimilarities between audio and visual modalities for deepfake detection. Similarly, MCL~\cite{liu2023mcl} utilizes contrastive loss between the cross-modal features and unimodal audio, frame, and video features to reduce cross-modal gaps and explore cues for multimodal deepfake detection. AVoiD-DF~\cite{yang2023avoid} introduces a Temporal-Spatial Encoder to extract audio-visual embeddings and a Multi-Modal Joint Decoder to jointly learn audio-visual inconsistency paired with contrastive loss for multimodal deepfake detection. MRDF~\cite{zou2024cross} introduces modality-specific and cross-modality regularization within AV-Hubert~\cite{shi2022learning} to mitigate inconsistencies in multimodal representations caused by modality-specific manipulations. The model utilizes the modality-regularization term to help the features learn unimodal cues. However, they do not explicitly leverage the unimodal embeddings during inference. AV-DFD~\cite{zhou2021joint} proposes a two-plus-one stream model that utilizes cross-attention along the temporal dimension to exploit intrinsic audio-visual synchronization. However, single-stage supervised training may not fully leverage audio-visual alignment, reducing its effectiveness against unseen deepfakes. To address this, recent studies~\cite{yu2023pvass, cheng2023voice} highlight the benefits of multimodal self-supervised pretraining.~\cite{cheng2023voice} pre-trains their VFD model using contrastive loss to exploit voice-face matching before fine-tuning it for deepfake classification. AVAD~\cite{feng2023self} leverages a synchronization model from~\cite{chen2021audio} to pretrain on audio-visual temporal synchronization, using the inferred features for fully unsupervised multimodal deepfake detection. Oorloff et al. propose AVFF~\cite{oorloff2024avff}, which pre-trains a Masked Autoencoder (MAE) framework inspired by CAV-MAE~\cite{gong2022contrastive} for joint audio-visual representation learning, later fine-tuned for multimodal deepfake classification. While AVAD~\cite{feng2023self} and AVFF~\cite{oorloff2024avff} achieve effective generalization against unseen manipulation or data distributions, they are two-stage approaches that require pretraining. Additionally, they primarily focus on cross-modal correspondence, which may cause them to overlook uni-modal inconsistencies. As a result, their effectiveness could be limited when detecting deepfakes that maintain audio-visual alignment. Moreover, most of these approaches do not explore the adaptability to Deepfake Temporal Localization.

\subsection{Temporal Deepfake Localization}
Temporal deepfake localization is a specialized form of temporal action detection, where the only "action" is identifying deepfake segments. One approach, boundary prediction, directly estimates temporal segments using boundary confidence~\cite{lin2018bsn, lin2019bmn, su2021bsn++}. BMN~\cite{lin2019bmn} and BSN++~\cite{su2021bsn++} generate a two-dimensional boundary-matching confidence map to represent each candidate proposal's start, duration, and confidence score. ActionFormer~\cite{zhang2022actionformer} enhances boundary regression by capturing hierarchical temporal features at multiple resolutions. Tridet~\cite{shi2023tridet} further improves ActionFormer by replacing its standard transformer block with the proposed SGP block.

For audio-visual deepfake temporal localization, BA-TFD~\cite{cai2022you} applies BMN~\cite{lin2019bmn} to predict boundary-matching confidence maps for each modality separately before fusing them for final proposal generation. BA-TFD+\cite{cai2023glitch} adopts the same architecture but replaces BMN’s boundary map prediction module with BSN++\cite{su2021bsn++}. The first multimodal deepfake temporal localization dataset, LAV-DF, was introduced in~\cite{chugh2020not, cai2023glitch}, followed by AV-Deepfake1M~\cite{cai2023av}. UMMAFormer~\cite{zhang2023ummaformer}, inspired by ActionFormer’s hierarchical processing, introduces a Parallel Cross-Attention Feature Pyramid Network (PCA-FPN) for deepfake temporal localization. They evaluate their model on the LAV-DF dataset and their proposed Temporal Video Inpainting Localization (TVIL) dataset.
\section{Proposed Method}
\label{sec:method}

\begin{figure*}
    \centering
    \includegraphics[width=\linewidth]{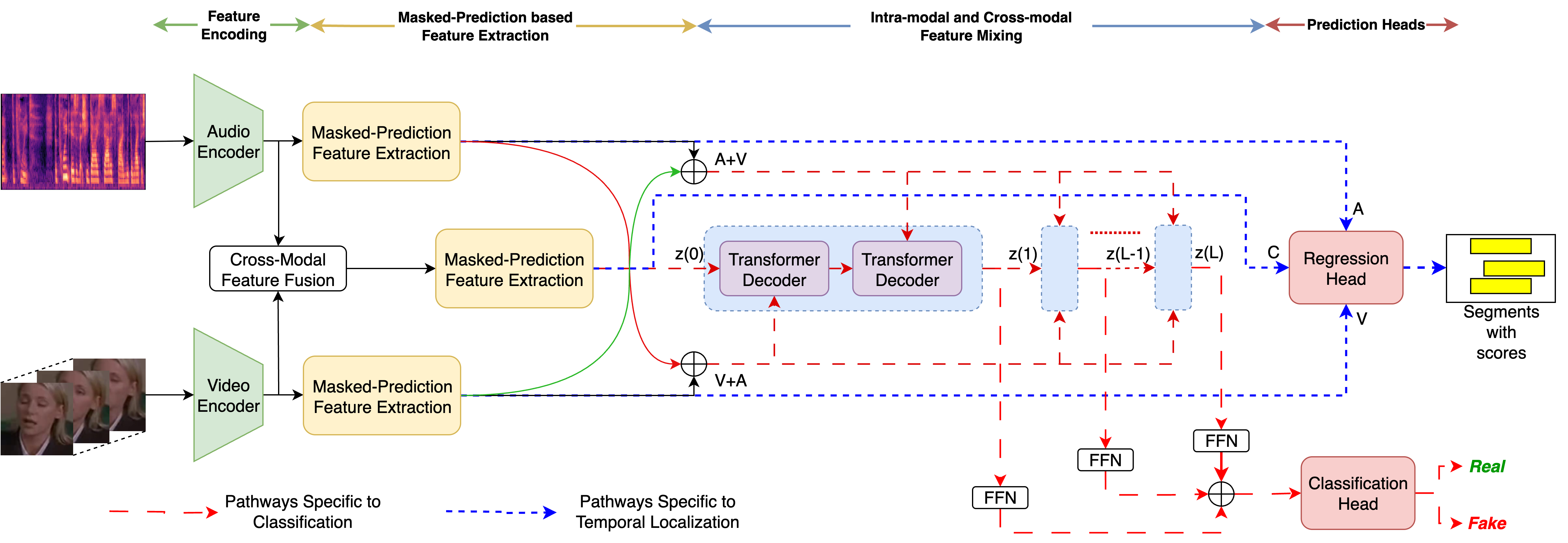}
    \caption{\textbf{Proposed Pipeline:} We extract unimodal embeddings and fuse them to create cross-modal features. Three masked-prediction modules detect intra-modal and cross-modal inconsistencies by predicting next-frame features and capturing deviations between predicted and actual features. We then fuse the intra-modal and cross-modal features through alternating cross-attention layers, and finally use the combined output for deepfake detection or temporal localization.}
    \label{fig:classification}
\end{figure*}

The proposed model consists of four key modules trained end-to-end in a single stage: (i) Feature Encoding, (ii) Masked-Prediction-Based Feature Extraction, (iii) Intra-modal and Cross-modal Feature Mixing, and (iv) Prediction Head (for classification or regression).  

The Feature Encoding module extracts frame-level unimodal embeddings using modality-specific encoders and fuses them to create cross-modal features. The Masked-Prediction-Based Feature Extraction module then detects intra-modal and cross-modal inconsistencies in the video using three masked-prediction modules. Each module learns frame-level deviations from real videos, guided by contrastive loss (see figure \ref{fig:classification} and figure \ref{fig:masked_pred}).  

Each masked-prediction module processes unimodal or cross-modal embeddings, encoding past temporal information into frame-level features using a causal transformer encoder. A causal transformer decoder then predicts the next frame’s features using only past information. A local-window-based attention mechanism computes deviations between the predicted and actual frame-level features.  

The Intra-modal and Cross-modal Feature Mixing module integrates the outputs of the three masked-prediction modules, enabling the model to detect inconsistencies both within and across modalities. Finally, the Prediction Head (for classification or regression) utilizes the combined output for deepfake detection or temporal localization.

\subsection{Feature Encoding}
We represent a sample input with a visual component $x_v \in R^{T_v \times C \times H \times W} $, where $ T_v, C, H,$ and $W$ denote the number of visual frames, channels, height, and width, and an audio component $x_a \in R^{T_a \times B}$, where $T_a$ and $B$ represent the number of audio frames and mel-bins, respectively. Each component is processed by its encoder, producing unimodal features $a$ and $v \in R^{T_v \times f}$, where $f$ is the feature dimension. Since our method is based on next-frame prediction, the encoders operate frame-wise, ensuring that frame-level features contain no information from other frames.

\subsubsection{Cross-Modal Feature Fusion}

\begin{figure}
    \centering
    \includegraphics[width=0.9\linewidth]{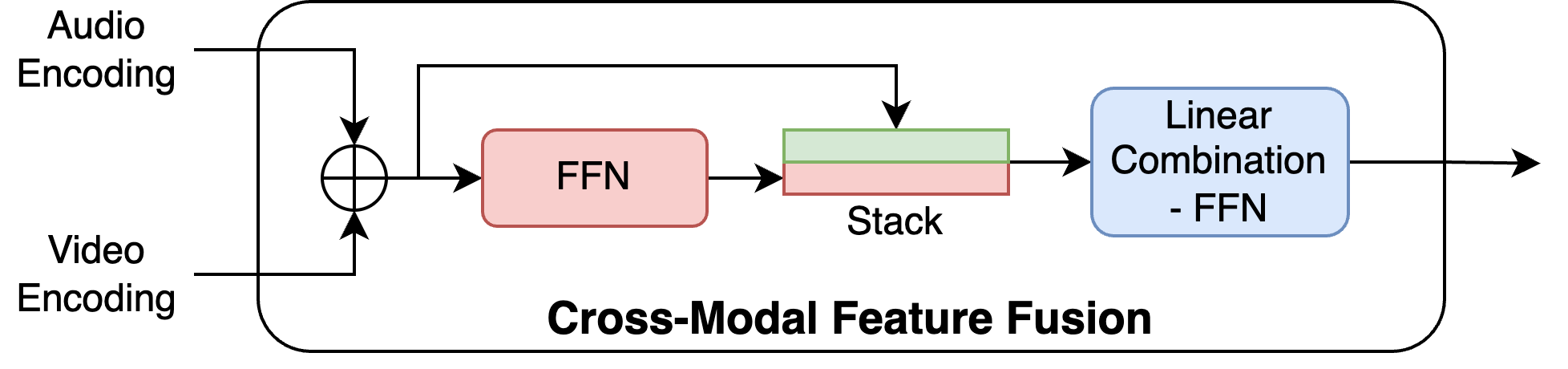}
    \caption{\textbf{Cross-Modal Feature Fusion:} Cross-modal features are formed by concatenating visual and audio encodings, processed through linear layers to learn a refined fusion representation.}
    \label{fig:cross_fusion}
\end{figure}

As shown in figure \ref{fig:cross_fusion}, we first concatenate the visual and audio encodings $v$ and $a$ to form the cross-modal feature, $c' \in R^{T_v \times 2f}$. This is then processed by a linear layer for feature mixing along the channel dimension. Finally, the cross-modal encoding $c \in R^{T_v \times 2f}$ is obtained as a learned linear combination of the linear layer’s output and $c'$, with trainable weights and biases. In other words, we stack the linear layer's output and $c'$ and pass the resulting embedding, $s \in R^{T_v \times 2f \times 2}$, through another linear layer to obtain $c$. The stacking enables the model to adaptively balance $c'$ and the linear layer output through learnable weights.

\subsection{Masked-Prediction based Feature Extraction}
\label{subsec:masked-feature}
\begin{figure*}
    \centering
    \includegraphics[width=0.8\linewidth]{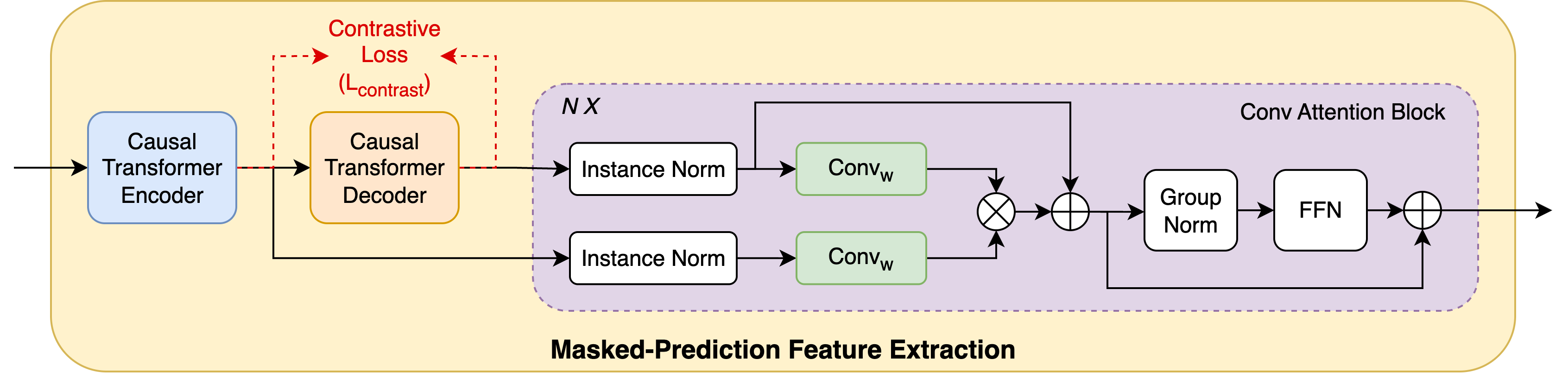}
    \caption{\textbf{Masked-Prediction based Feature Extraction:} We extract intra-modal and cross-modal inconsistencies by measuring deviations between predicted and actual frame-level features. Local convolution-based attention is applied to detect inconsistencies. To enhance adaptability for both classification and localization, frame-level contrastive loss is applied, ensuring the model learns to distinguish real and manipulated frames effectively.}
    \label{fig:masked_pred}
\end{figure*}

This stage focuses on extracting features that highlight local intra-modal and cross-modal inconsistencies around each frame in the input video. To achieve this, the model measures deviations between predicted and actual frame-level features across both modalities. While prior works ~\cite{oorloff2024avff, feng2023self} show that audio-visual correspondence aids multimodal deepfake detection, ~\cite{feng2023self} notes that such models struggle against attacks that maintain audio-visual consistency throughout the entire length of the video. To address this, we train three masked-prediction modules to capture both intra-modal and cross-modal inconsistencies. The dimensions for intra-modal features remain $T_v \times f$, while those for cross-modal features are $T_v \times 2f$.  Additionally, to detect inconsistencies, we apply attention between predicted and actual features. Since we need to focus on a local region around each frame, we develop a local convolution-based attention mechanism. Finally, to ensure adaptability to both classification and localization (where deepfakes may appear in only some frames), we guide our masked-prediction modules using frame-level contrastive loss. The overall pipeline is illustrated in figure \ref{fig:masked_pred}.

\subsubsection{Causal Transformer Encoder}
\label{sec:causal_trans_enc}
The proposed sub-module processes frame-level features using a transformer encoder with a causal mask. This serves two purposes: (i) it enables effective feature mixing, particularly in cross-modal extraction, by combining audio and visual features into a joint representation; and (ii) it integrates information from previous frames to capture temporal dependencies. The causal mask ensures the model only considers past frames, preventing leakage from future frames. As a result, the output contains well-mixed frame-level representations enriched with past context.

\subsubsection{Causal Transformer Decoder}
This sub-module predicts the next frame's features by analyzing previous ones. We pass the encoder's output through a transformer decoder, applying a causal mask to prevent the model from accessing future frames. The decoder's frame-level output predicts the features of the next frame. To align this with the input, we shift the output by one frame and append the first frame of the causal encoder's output to the first frame of the decoder's output. The shifted feature is then forwarded as the module's output for further processing.

\subsubsection{Convolutional Cross-Attention}
\label{sec:conv_cross_attend}
The sub-module captures the difference between predicted and actual frame-level features, which are outputs from the next-frame predictor and causal encoder, respectively. 
A pointwise distance is a simple way to measure differences between two features, but it treats all values in the feature matrix equally. Instead, we use cross-attention, allowing the model to learn adaptive weights based on feature similarity. A typical transformer's cross-attention layer attends to all frames at once, but this results in high computational complexity due to dense pairwise calculations. Since our framework requires attention only around each frame, we use local window attention. Inspired by ~\cite{shi2023tridet}, we replace regular cross-attention with convolutional window-based attention. The $Conv_w$ in figure \ref{fig:masked_pred} represents a depthwise convolution with kernel size $w$. To focus on local attention, we set $w=9$ after conducting ablation over different values (Section~\ref{subsubsec:hyperparam}). Additionally, following ~\cite{shi2023tridet}, we replace LayerNorm ~\cite{ba2016layer} with GroupNorm ~\cite{wu2018group}. The output of the convolutional attention block is represented by $C \in R^{T_v \times 2f}$ for cross-modal, $A \in R^{T_v \times f}$ for audio, and $V \in R^{T_v \times f}$ for video modality.

\subsection{Intra-modal and Cross-modal Feature Mixing}

As shown in figure \ref{fig:classification}, the proposed pipeline enhances the cross-modal output $C$ from the masked-frame prediction module, also denoted as $z(0) \in R^{T_v \times 2f}$, by incorporating the visual and audio outputs $A$ and $V \in R^{T_v \times f}$. This is achieved through $L$ levels of attention, each containing two transformer decoders. Since $z(0)$ has twice the feature dimension of $A$ and $V$, we create two encodings by concatenating $A$ and $V$ in different orders, labeled $A+V$ and $V+A$. Each attention level applies cross-attention between the previous level's output $z(l - 1)$ and $V+A$, followed by another cross-attention with $A+V$. This alternating cross-attention ensures that each channel in $z(l - 1)$ attends to both $A$ and $V$, effectively capturing both cross-modal and intra-modal inconsistencies. The output of the feature mixing module is then used by the prediction head for classification.

\subsection{Deepfake Classification}

For detection, the model processes the outputs of the feature mixing module, $z(l)$, at each level $l$. These are passed through a linear layer to produce features $x(l) \in R^{T_v \times f}$, which are then concatenated along the channel dimension and permuted. The resulting feature $x \in R^{Lf \times T_v}$ is fed into the classification head for prediction. The classification head consists of a linear layer followed by four 1D convolutional blocks with output channels $[2f, f, f/2, f/4]$, kernel size 3 and stride 2 in sequence, ReLU activations, a flatten function, and two linear layers with output dimensions 128 and 1, respectively, producing the final logits.

\subsection{Deepfake Temporal Localization}

\begin{figure*}
    \centering
    \includegraphics[width=0.9\linewidth]{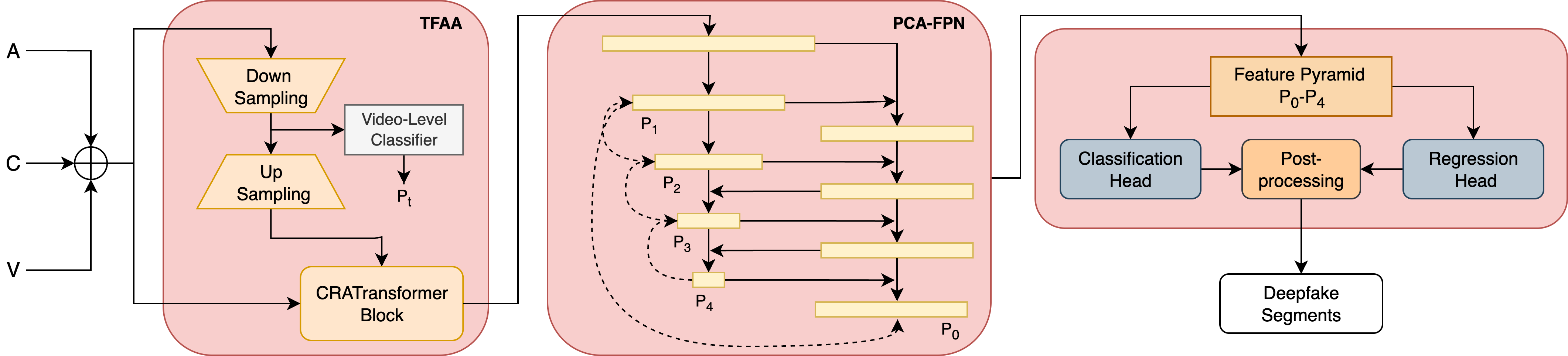}
    \caption{\textbf{Regression Head:} We extract the outputs from the intra-modal and cross-modal Masked-Prediction Feature Extraction modules, denoted by A, V, and C. We then concatenate these three features along the feature dimension and pass them through the adapted UMMAFormer~\cite{zhang2023ummaformer} model.}
    \label{fig:reghead}
\end{figure*}

Figures \ref{fig:classification} and \ref{fig:reghead} present the full localization pipeline. This pipeline follows the same procedure as classification up to the Masked-Prediction Feature Extraction stage. The outputs of the three Masked-Prediction Feature Extraction modules, $A$, $V$, and $C$, are concatenated along the feature dimension and directly passed to the regression head. Our regression head is based on the state-of-the-art UMMAFormer~\cite{zhang2023ummaformer}. Unlike the classification branch, we omit the Intra-modal and Cross-modal Feature Mixing before the regression stage, since UMMAFormer itself incorporates multi-level feature enhancement. Specifically, the UMMAFormer head first applies an encoder–decoder reconstruction, followed by a transformer block that takes both the original and reconstructed features as input. The output is then processed by a Parallel Cross-Attention Feature Pyramid Network (PCA-FPN), inspired by ActionFormer~\cite{zhang2022actionformer}, which hierarchically downsamples the sequence into five temporal levels. This feature pyramid is finally used to produce frame-level deepfake predictions and to estimate the start and end timestamps of the nearest manipulated segment at each resolution level.

\subsection{Loss Functions}

For classification, we compute the predicted logits and optimize them using binary cross-entropy loss against the ground truth labels. For localization, we adopt all the losses from the UMMAFormer~\cite{zhang2023ummaformer} model.  

Additionally, we guide the masked-prediction feature extraction module to accurately predict the next frame’s features. Since the dataset contains both real and deepfake samples, we employ contrastive loss to bring predicted and actual frame features closer for real samples while pushing them apart for deepfakes. This helps the model learn distinguishing features, improving the convolutional attention block's effectiveness. As we obtain three contrastive loss values (two intra-modal and one cross-modal), we average them before incorporating them into the classification or regression losses.

Hence, the overall classification loss, $\mathcal{L}_{cls}$, can be given by

\begin{equation}
    \mathcal{L}_{cls} = \mathcal{L}_{BCE} + \frac{\mathcal{L}_{contrast}^{a} + \mathcal{L}_{contrast}^{v} + \mathcal{L}_{contrast}^{av}}{3},
    \label{eq:Loss_Class}
\end{equation}

where $\mathcal{L}_{BCE}$ is the binary cross-entropy loss, and $\mathcal{L}_{contrast}^{a}$, $\mathcal{L}_{contrast}^{v},$ and $\mathcal{L}_{contrast}^{av}$ are the contrastive loss calculated for audio, visual and audio-visual modalities respectively.

Similarly, the overall localization loss, $\mathcal{L}_{reg}$, can be given by

\begin{multline}
    \mathcal{L}_{reg} = \mathcal{L}_{cls}^{U} + \lambda_{reg}^{U}\mathcal{L}_{reg}^{U} + \\ \lambda_{rec}^{U}\mathcal{L}_{rec}^{U} + \lambda_{scls}^{U}\mathcal{L}_{scls}^{U} + \\ \frac{\mathcal{L}_{contrast}^{a} + \mathcal{L}_{contrast}^{v} + \mathcal{L}_{contrast}^{av}}{3},
    \label{eq:Loss_Reg}
\end{multline}

where$\mathcal{L}_{cls}^{U}$ and $\mathcal{L}_{reg}^{U}$ denote the classification and regression losses adopted from the UMMAFormer~\cite{zhang2023ummaformer} heads, while $\mathcal{L}_{rec}^{U}$ and $\mathcal{L}_{scls}^{U}$ represent the reconstruction loss and video-level focal loss, respectively, also taken from UMMAFormer. The hyperparameters are set as $\lambda_{reg}^{U}=2$, $\lambda_{rec}^{U}=1$, and $\lambda_{scls}^{U}=0.1$, following UMMAFormer.
\section{Experiments}
\label{sec:experiment}

\subsection{Implementation Details}

The videos are extracted at 25 fps, and the audio is sampled at 16 kHz. The audio is converted into a mel spectrogram using 64 mel filter banks, a window size of 321, and a hop length of 160.  

To ensure the encoder models remain lightweight and simple, we use AV-HuBERT's~\cite{shi2022learning} ResNet-18 visual encoder with 512 channels and a ViT~\cite{dosovitskiy2020image} audio encoder with a hidden size of 192, an MLP size of 768, 12 layers, and 3 heads. We modified the encoders to restrict its focus to a single frame. The outputs of both encoders are projected into a 256-dimensional space to align the visual and audio feature dimensions ($ f = 256 $).  

The Masked Prediction-based Feature Extraction module consists of three transformer encoder layers in the Causal Transformer Encoder and three transformer decoder layers in the Causal Transformer Decoder. Each encoder and decoder layer uses four attention heads with a feature dimension of 512. The module also includes three Convolution Attention Blocks ($N=3$).

The classification module contains three attention blocks ($ L = 3 $). The Regression head adopts all the same setup from UMMAFormer~\cite{zhang2023ummaformer}, except the input feature dimension, which is 1024 in our case.

\subsection{Datasets}

The classification network was trained on \textbf{FakeAVCeleb}~\cite{khalid2021fakeavceleb}. To balance the dataset, real video samples were augmented with samples from \textbf{VoxCeleb2}~\cite{Chung18b}. Cross-dataset generalization was assessed using a subset of \textbf{KoDF}~\cite{kwon2021kodf}. For training and evaluating the localization module, we used the \textbf{LavDF}~\cite{cai2023glitch, cai2022you} dataset. Details on each dataset are provided below.

\textbf{FakeAVCeleb~\cite{khalid2021fakeavceleb}}: The FakeAVCeleb dataset is developed for deepfake detection and consists of 20,000 video clips. It includes 500 authentic videos from VoxCeleb2~\cite{Chung18b} and 19,500 deepfake samples created using various manipulation techniques, including FaceswapGAN~\cite{nirkin2019fsgan}, Faceswap~\cite{korshunova2017fast}, SV2TTS~\cite{jia2018transfer}, and Wav2Lip~\cite{prajwal2020lip}. The dataset is categorized into four groups: Real Visuals - Real Audio, Fake Visuals - Real Audio, Fake Visuals - Fake Audio, and Real Visuals - Fake Audio.

\textbf{VoxCeleb2~\cite{Chung18b}}: VoxCeleb2 consists of real YouTube videos featuring over 6,000 celebrities, with more than 1 million utterances in the development set and approximately 36,000 in the test set. Each video includes interviews of celebrities with various ethnicities, accents, professions, and age groups. To address the imbalance in the FakeAVCeleb~\cite{khalid2021fakeavceleb} dataset, samples from VoxCeleb2 were used to augment the real videos during training.

\textbf{KoDF~\cite{kwon2021kodf}}: KoDF is a large-scale deepfake video dataset featuring 403 Korean subjects. It includes over 62,000 real videos and more than 175,000 deepfakes created using six different synthesis techniques. The dataset is used to evaluate cross-dataset generalization.

\textbf{LavDF~\cite{cai2022you, cai2023glitch}}: The LAV-DF dataset contains deepfakes where only specific segments of a video are altered. Like FakeAVCeleb, it includes approximately 36,000 real samples from VoxCeleb2, with manipulations applied to one or both modalities, resulting in over 99,000 deepfake samples. It features 114,253 fake segments ranging from 0 to 1.6 seconds in duration, with an average length of 0.65 seconds. Notably, 89.26\% of fake segments are shorter than 1 second. Videos in the dataset have a maximum duration of 20 seconds, with 69.61\% being under 10 seconds. The dataset is evenly divided into four categories: real, audio-modified, visual-modified, and both audio-visual-modified.

\subsection{Evaluation}

Following prior work~\cite{feng2023self, oorloff2024avff}, we evaluate our classification model against audio-visual and visual-only baselines under three settings: (i) intra-dataset, (ii) cross-manipulation, and (iii) cross-dataset generalization, using accuracy (ACC), average precision (AP), and AUC~\cite{haliassos2021lips, haliassos2022leveraging}. For temporal deepfake localization, we adopt the evaluation protocols of~\cite{cai2022you, cai2023glitch, zhang2023ummaformer} and other temporal action localization methods~\cite{shi2023tridet, zhang2022actionformer}, reporting Average Precision at IoU thresholds of 0.5, 0.75, and 0.95~\cite{caba2015activitynet}. In addition, for the localization ablation study, we also report mean Average Precision (mAP), computed as the mean AP across the three thresholds.

\subsubsection{Intra-Dataset Performance}

\begin{table}[ht]
\centering
\caption{\textbf{Intra-Dataset Result on FakeAVCeleb:} We divide the FakeAVCeleb dataset into a 70\%-30\% split for training and evaluation. Best results are highlighted in bold. AVFF fine-tunes a pretrained network during classification, adding extra parameters.}
\label{tab:intra-dataset}
    \begin{tabular}{r c c c}
        \toprule
        Method & Modality & ACC & AUC \\
        \midrule
        Xception~\cite{rossler2019faceforensics++} & \textcolor{orange}{V} & 67.9 & 70.5 \\
        LipForensics~\cite{haliassos2021lips}  & \textcolor{orange}{V} & 80.1 & 82.4 \\
        FTCN~\cite{zheng2021exploring} & \textcolor{orange}{V} & 64.9 & 84.0 \\
        CViT~\cite{wodajo2021deepfake} & \textcolor{orange}{V} & 69.7 & 71.8 \\
        RealForensics~\cite{haliassos2022leveraging} & \textcolor{orange}{V} & 89.9 & 94.6 \\
        \midrule
        VFD~\cite{cheng2023voice} & \textcolor{green}{AV} & 81.5 & 86.1 \\
        AVoid-DF~\cite{yang2023avoid} & \textcolor{green}{AV} & 83.7 & 89.2 \\
        MCL~\cite{liu2023mcl} & \textcolor{green}{AV} & 85.9 & 89.2 \\
        Multimodaltrace~\cite{raza2023multimodaltrace} & \textcolor{green}{AV} & 92.9 & - \\
        MRDF-Margin~\cite{zou2024cross} & \textcolor{green}{AV} & 93.4 & 91.8 \\
        MRDF-CE~\cite{zou2024cross} & \textcolor{green}{AV} & 94.1 & 92.4 \\
        PVASS~\cite{yu2023pvass} & \textcolor{green}{AV} & 95.7 & 97.3 \\
        AVFF~\cite{oorloff2024avff} & \textcolor{green}{AV} & \textbf{98.6} & \textbf{99.1} \\
        \midrule  
        Ours & \textcolor{green}{AV} & 94.8 & 96.3 \\
        \bottomrule
    \end{tabular}
\end{table}

Following~\cite{oorloff2024avff,yang2023avoid}, we train our model on 70\% of the FakeAVCeleb dataset and evaluate it on the remaining 30\%, with results summarized in Table~\ref{tab:intra-dataset}. Our approach achieves substantial gains over prior work, improving upon Avoid-DF~\cite{yang2023avoid} by 11.1\% in accuracy and 7.1\% in AUC, while performing competitively with the state-of-the-art AVFF~\cite{oorloff2024avff}. Compared to visual-only models, the advantage of audio-visual methods is clear, as they capture cross-modal consistencies rather than relying solely on visual artifacts. Our model outperforms most audio-visual methods. PVASS surpasses our model and AVFF achieves state-of-the-art results, but they both depend on pre-training, and AVFF requires additional fine-tuning during classification. In contrast, our single-stage framework attains comparable performance without this additional complexity. Importantly, relative to MRDF~\cite{zou2024cross}, which also employs a single-stage design and considers both intra-modal and cross-modal cues during training, our model improves AUC by nearly 4\% over its MRDF-CE variant. This improvement stems from a key difference: MRDF discards unimodal heads at inference and relies solely on the cross-modal transformer, while our method explicitly models both unimodal and cross-modal cues and integrates them through cross-attention, allowing inconsistencies to be directly exploited for prediction.

\subsubsection{Cross-Manipulation Generalization}

\begin{table*}
    \centering
    \caption{\textbf{Cross-Manipulation Detection on FakeAVCeleb:} We report the Average Precision(AP) and AUC scores over one manipulation category while training on the rest of the data. For this, we consider five categories: RVFA, FVRA-WL, FVFA-FS, FVFA-WL, and FVFA-GAN. The column AVG-FV refers to the mean of the performance achieved on the four Fake Visual categories. Bold highlights the best performance for every metric under each category.}
    \label{tab:cross-manipulations}
    \begin{tabular}{r c c c c c c c c c c c c c}
        \toprule
        Method & Modality & \multicolumn{2}{c}{RVFA} & \multicolumn{2}{c}{FVRA-WL} & \multicolumn{2}{c}{FVFA-FS} & \multicolumn{2}{c}{FVFA-WL} & \multicolumn{2}{c}{FVFA-GAN} & \multicolumn{2}{c}{AVG-FV} \\
        \cmidrule(lr){3-4} \cmidrule(lr){5-6} \cmidrule(lr){7-8} \cmidrule(lr){9-10} \cmidrule(lr){11-12} \cmidrule(lr){13-14}
        & & AP & AUC & AP & AUC & AP & AUC & AP & AUC & AP & AUC & AP & AUC \\
        \midrule
        Xception~\cite{rossler2019faceforensics++}          & \textcolor{orange}{V}  & -    & - & 88.2 & 88.3   & 92.3 & 93.5 & 91.0 & 91.0 & 67.6 & 68.5 & 84.8 & 85.3 \\
        LipForensics~\cite{haliassos2021lips}       & \textcolor{orange}{V}  & -    & -  & 97.8 & 97.7  & 99.9 & 99.9 & 98.6 & 98.7 & 61.5 & 68.1 & 89.4 & 91.1 \\
        FTCN~\cite{zheng2021exploring}              & \textcolor{orange}{V}  & -    & -  & 96.2 & 97.4   & \textbf{100.} & \textbf{100.} & 95.6 & 96.5 & 77.4 & 78.3 & 92.3 & 93.1 \\
        RealForensics~\cite{haliassos2022leveraging}      & \textcolor{orange}{V}  & -    & - & 88.8 & 93.0   & 99.3 & 99.1 & 93.4 & 96.7 & 99.8 & 99.8 & 95.3 & 97.1 \\
        \midrule
        AV-DFD~\cite{zhou2021joint}            & \textcolor{green}{AV}  & 74.9 & 73.3 & 97.0 & 97.4 & 99.6 & 99.7 & \textbf{100.} & \textbf{100.} & 58.4 & 55.4 & 88.8 & 88.1 \\
        AVAD (LRS2)~\cite{feng2023self}       & \textcolor{green}{AV}  & 62.4 & 71.6 & 93.6 & 93.7 & 95.3 & 95.8 & 93.8 & 94.1 & 94.1 & 94.1 & 94.2 & 94.5 \\
        AVAD (LRS3)~\cite{feng2023self}       & \textcolor{green}{AV}  & 70.7 & 80.5 & 91.1 & 93.0 & 91.0 & 92.3 & 91.4 & 93.1 & 91.6 & 92.7 & 91.3 & 92.8 \\
        AVFF~\cite{oorloff2024avff}           & \textcolor{green}{AV}  & 93.3 & 92.4 & 94.8 & \textbf{98.2} & \textbf{100.} & \textbf{100.} & 99.4 & 99.8 & \textbf{99.9} & \textbf{100.} & 98.5 & \textbf{99.5} \\
        \midrule
        Ours  & \textcolor{green}{AV} & \textbf{96.4} & \textbf{95.9} & \textbf{98.5} & 88.6 & 99.7 & 98.4 & 99.7 & 97.6 & 99.7 & 97.8 & \textbf{99.4} & 95.6 \\
        \bottomrule
    \end{tabular}
\end{table*}

We adopt an evaluation strategy similar to~\cite{feng2023self, oorloff2024avff} to evaluate the model's generalization to unseen manipulation techniques. We divide the FakeAVCeleb dataset into five categories: (i) \textbf{RVFA:} Real Visual - Fake Audio generated using SV2TTS~\cite{jia2018transfer},(ii) \textbf{FVRA-WL:} Fake Visual - Real Audio generated using Wav2Lip~\cite{prajwal2020lip},(iii) \textbf{FVFA-FS:} Fake Visual - Fake Audio generated using FaceSwap~\cite{korshunova2017fast}, Wav2Lip~\cite{prajwal2020lip} and SV2TTS~\cite{jia2018transfer}), (iv) \textbf{FVFA-WL:} Fake Visual - Fake Audio generated by Wav2Lip~\cite{prajwal2020lip} and SV2TTS~\cite{jia2018transfer}), and (v) \textbf{FVFA-GAN:} Fake Visual - Fake Audio generated usinf FaceSwapGAN~\cite{nirkin2019fsgan}, Wav2Lip~\cite{prajwal2020lip} and SV2TTS~\cite{jia2018transfer}). We evaluate our model using a leave-one-out approach, where we use test samples from one category to assess the model trained on the rest of the training dataset.

Table \ref{tab:cross-manipulations} shows that most visual-only models exhibit inconsistent performance across manipulation categories. While they perform well on FVFA-WL and FVFA-FS, their performance drops on FVFA-GAN. RealForensics~\cite{haliassos2022leveraging} achieves strong results across all fake video categories but requires pre-training and is limited to visual deepfakes. AVFF sets the state-of-the-art in most of the categories but, like RealForensics and AVAD~\cite{feng2023self}, relies on pre-training and additional fine-tuning during classification. In contrast, our single-stage model delivers performance comparable to state-of-the-art methods. It also shows a significant performance boost under RVFA, similar to AVFF~\cite{oorloff2024avff}, due to its consistency across manipulation categories and even sets up a new state-of-the-art performance. Compared to AV-DFD~\cite{zhou2021joint}, another single-stage multimodal deepfake detector, our model improves AP by 21.5\% and AUC by 22.6\% on RVFA and AP by 41.3\% and AUC by 42.4\% on FVFA-GAN, demonstrating its robustness and reliability.

\subsubsection{Cross-Dataset Generalization}
\begin{table}
    \centering
    \caption{\textbf{Cross-Dataset performance on KoDF:} AP and AUC score(\%) achieved on KoDF dataset by the models trained on FakeAVCeleb. The best results are highlighted in bold.}
    \label{tab:cross-dataset} 
    \begin{tabular}{r c c c }
        \toprule
        Method & Modality & \multicolumn{2}{c}{KoDF~\cite{kwon2021kodf}} \\
        \cmidrule(lr){3-4}
        & & AP & AUC \\
        \midrule
        Xception~\cite{rossler2019faceforensics++}  & \textcolor{orange}{V}  & 76.9    & 77.7 \\
        LipForensics~\cite{haliassos2021lips}       & \textcolor{orange}{V}  & 89.5    & 86.6  \\
        FTCN~\cite{zheng2021exploring}              & \textcolor{orange}{V}  & 66.8    & 68.1 \\
        RealForensics~\cite{haliassos2022leveraging}& \textcolor{orange}{V}  & 95.7    & 93.6 \\
        \midrule
        AV-DFD~\cite{zhou2021joint}                 & \textcolor{green}{AV}  & 79.6 & 82.1 \\
        AVAD~\cite{feng2023self}                    & \textcolor{green}{AV}  & 87.6 & 86.9 \\
        AVFF~\cite{oorloff2024avff}                 & \textcolor{green}{AV}  & 93.1 & 95.5 \\
        \midrule
        Ours                                        & \textcolor{green}{AV} & \textbf{100} & \textbf{100} \\
        \bottomrule
    \end{tabular}
\end{table}
To evaluate our model’s effectiveness across different data distributions, we tested it on a subset of the KoDF dataset~\cite{kwon2021kodf}, following the protocol in~\cite{feng2023self, oorloff2024avff}. Since KoDF videos are much longer than those in FakeAVCeleb, we applied max pooling to the outputs of the Intra-modal and Cross-modal Feature Mixing step along the temporal dimension before feeding them to the final classification head. For strict evaluation, we used a model trained on FakeAVCeleb samples excluding those from the FVRA-WL category, as KoDF contains FVRA samples generated with Wav2Lip. Table~\ref{tab:cross-dataset} shows the results, where our model outperforms existing approaches and achieves perfect scores. Notably, despite being trained only on FakeAVCeleb, which contains English videos, our approach generalizes well to the Korean KoDF dataset.

Since the model achieves perfect scores on the KoDF dataset, we further tested its generalization capability on the CREMA dataset~\cite{stypulkowski2024diffused}, following~\cite{kamat2023revisiting}, which contains 820 deepfake samples generated using recent diffusion models. We augmented the test set with 820 real samples from the VoxCeleb2 dataset~\cite{Chung18b} and achieved an AP of 97.65 and an AUC of 98.64, demonstrating strong generalization to diffusion-generated deepfakes as well.

\subsubsection{Temporal Deepfake Localization}

We evaluate our localization module on the LAV-DF~\cite{cai2023glitch, cai2022you} dataset, with results shown in Table \ref{tab:lavdf_result}. Our model surpasses existing methods and sets a new state-of-the-art across all IoU thresholds, achieving a 19.82\% AP gain at 95\% IoU over UMMAFormer~\cite{zhang2023ummaformer}. Visual-only methods such as MDS~\cite{chugh2020not} and BMN~\cite{lin2019bmn} perform poorly on LAV-DF, which contains both unimodal and multimodal deepfakes. Multimodal approaches~\cite{bagchi2021hear, cai2022you} perform well at 0.5 IoU but drop by nearly 40\% at 0.75 IoU due to weak feature enhancement for short deepfake segments~\cite{zhang2023ummaformer}. To mitigate this, recent methods~\cite{zhang2022actionformer, shi2023tridet, zhang2023ummaformer} use hierarchical feature processing to sustain performance at higher IoU thresholds. Our model maintains strong results across all IoU levels, and the large improvement at 95\% IoU over UMMAFormer highlights the effectiveness of our masked-prediction-based feature extraction in detecting short deepfake segments.
\begin{table}
    \centering
    \caption{\textbf{Temporal Localization on Lav-DF:} We report Average Precision(\%) achieved at an IOU threshold of 0.5, 0.75, and 0.95 on the LAV-DF dataset. The best scores are highlighted in bold.}
    \label{tab:lavdf_result}
    \begin{tabular}{r c c c}
        \toprule
        Method & AP@0.5 & AP@0.75 & AP@0.95 \\
        \midrule
        MDS~\cite{chugh2020not} & 12.78 & 1.62 & 0.00 \\
        BMN~\cite{lin2019bmn} & 24.01 & 7.61 & 0.07 \\
        AVFusion~\cite{bagchi2021hear} & 65.38 & 23.89 & 0.11 \\
        BA-TFD~\cite{cai2022you} & 76.90 & 38.50 & 0.25 \\
        ActionFormer~\cite{zhang2022actionformer} & 85.23 & 59.05 & 00.93 \\
        TriDet~\cite{shi2023tridet} & 86.33 & 70.23 & 03.05 \\
        BA-TFD+~\cite{cai2023glitch} & 96.30 & 84.96 & 04.44 \\
        UMMAFormer~\cite{zhang2023ummaformer} & 98.83 & 95.54 & 37.61 \\
        \midrule
        Ours  & \textbf{99.25} & \textbf{97.40} & \textbf{57.43} \\
        \bottomrule
    \end{tabular}
\end{table}

\subsection{Ablation Studies}

\begin{table*}
    \centering
    \caption{\textbf{Hyperparameter Ablation:} We experimented with different Kernel Sizes ($w$), Depths ($N$) and replacing the Convolution Attention with Transformer Decoder. We report the performance of each setup on the LAV-DF dataset in terms of Average Precision at the IoU thresholds of 50, 75, and 95\%. Additionally, we also report the mean of the three average precisions (mAP). The rows denoting the selected hyperparameter value are highlighted in bold. The best values under each metric are underlined.}
    \label{tab:Hyperparameter}
    \begin{tabular}{c c c c c c c}
        \toprule
        Kernel Size & Depth       & Attention               & AP@0.5            & AP@0.75        & AP@0.95        & mAP \\
        \midrule
        1           & 3           & Convolution             & 99.25             & 97.16          & 55.14          & 83.85 \\
        3           & 3           & Convolution             & 98.99             & 97.03          & 56.25          & 84.09 \\
        5           & 3           & Convolution             & 98.12             & 95.78          & 57.89          & 83.93 \\
        7           & 3           & Convolution             & \underline{99.26} & \underline{97.47} & 54.88       & 83.87 \\
        \textbf{9}  & \textbf{3}  & \textbf{Convolution}    & \textbf{99.25}    & \textbf{97.40} & \textbf{57.43} & \underline{\textbf{84.69}} \\
        11          & 3           & Convolution             & 98.72             & 96.63          & \underline{58.22} & 84.52 \\
        13          & 3           & Convolution             & 98.97             & 97.07          & 55.14          & 83.73 \\
        15          & 3           & Convolution             & 98.87             & 96.79          & 54.33          & 83.33 \\
        \midrule
        9           & 2           & Convolution             & 98.50             & 96.14          & 53.47          & 82.70 \\
        \textbf{9}  & \textbf{3}  & \textbf{Convolution}    & \underline{\textbf{99.25}}    & \underline{\textbf{97.40}} & \underline{\textbf{57.43}} & \underline{\textbf{84.69}} \\
        9           & 4           & Convolution             & 98.42             & 96.08          & 57.01          & 83.84 \\
        9           & 5           & Convolution             & 99.06             & 97.22          & 55.88          & 84.05 \\
        \midrule
        -           & 1           & Transformer             & 98.84             & 96.63          & 50.07          & 81.85 \\
        -           & 3           & Transformer             & 04.52             & 00.78          & 00.00          & 01.77 \\
        \textbf{9}  & \textbf{3}  & \textbf{Convolution}    & \underline{\textbf{99.25}}    & \underline{\textbf{97.40}} & \underline{\textbf{57.43}} & \underline{\textbf{84.69}} \\
        \bottomrule
    \end{tabular}
\end{table*}

\begin{table}
    \centering
    \caption{\textbf{Feature Set Analysis and Impact of Contrastive loss:} We report the classification performance achieved by the model when trained on only cross-modal (C) or single-modality (A and V) features and without contrastive loss. The best results are highlighted in bold}
    \label{tab:feature_set}
    \begin{tabular}{l c c c}
        \toprule
        Feature set                   & Contrastive Loss & AP   & AUC \\
        \midrule
        Only C                        & \checkmark       & 99.3 & 77.5 \\
        Only V                        & \checkmark       & 99.8 & 92.2 \\
        Only A                        & \checkmark       & 99.3 & 76.3 \\
        \midrule
        V+A                           & \checkmark       & 99.5 & 82.5 \\
        \midrule
        C+V+A                         & \ding{55}        & 99.5 & 81.3 \\
        \midrule
        \textbf{Ours (C+V+A)}         & \checkmark       & \textbf{99.9} & \textbf{96.3} \\
        \bottomrule
    \end{tabular}
\end{table}

\subsubsection{Hyperparameter Ablations}
\label{subsubsec:hyperparam}

Since Convolutional Cross-Attention (Section~\ref{sec:conv_cross_attend}) is one of our main contributions and a key part of the Masked-Prediction-based Feature Extraction module, we ablate two major hyperparameters: (i) Kernel Size ($w$), which controls the window size of local attention around each frame, and (ii) Depth ($N$), which denotes the number of Conv Attention Blocks (Figure~\ref{fig:masked_pred}). Because our motivation for local attention is closely tied to temporal localization, we conduct ablations on LAV-DF and present the results in Table~\ref{tab:Hyperparameter}. Specifically, we test kernel sizes $w \in \{1, 3, 5, 7, 9, 11, 13, 15\}$ and depths $N \in \{2, 3, 4, 5\}$. For the depth experiments, we fix $w=9$, and for kernel size experiments, we fix $N=3$. The results show that depth $N=3$ consistently provides the best performance across all metrics, making it the clear choice. For kernel size, we observe a trade-off: as AP at 50\% and 75\% thresholds improves, AP at the stricter 95\% threshold often declines, and vice-versa. We therefore select $w=9$, which deviates minimally from the best value under each metric and yields the highest overall mean Average Precision (mAP), striking a balance between lenient (50, 75) and strict (95) thresholds. Very small kernel sizes fail to capture sufficient temporal context, while very large ones introduce excessive smoothing, potentially blurring sharp boundaries.

\subsubsection{Impact of Convolutional Attention}

To evaluate the impact of the convolutional attention block, designed to capture local attention around each frame, we replaced it with a transformer decoder and report the results in Table~\ref{tab:Hyperparameter}. We tested two configurations: (i) a 3-layer transformer decoder (depth 3) with 135M parameters compared to our 127M, and (ii) a 1-layer transformer decoder (depth 1) with 126M parameters, close to ours. The 3-layer decoder performed poorly, likely due to overfitting from its larger parameter size, while the 1-layer decoder outperformed UMMAFormer~\cite{zhang2023ummaformer} on all metrics. However, both setups were inferior to convolutional attention. In particular, Convolutional Attention improved performance by 7.36\% at the strict 95\% threshold. This is because Convolutional Attention emphasizes local temporal relations, preserving sharper boundaries, whereas the Transformer attends globally, which can blur the distinction between real and fake segments.

\subsubsection{Feature Set Analysis}

To assess the impact of individual cross-modal and intra-modal features, we trained the classification model using only one feature at a time instead of all three. Since this setup involves a single feature, alternating cross-attention was unnecessary, so we replaced the transformer decoder layers in the classification module with an equal number of transformer encoder layers. Additionally, only the contrastive loss corresponding to the selected feature was used during training. Table \ref{tab:feature_set} presents the results for the three individual feature sets: `C' (cross-modal features), `A' (audio features), and `V' (video features), or just a simple concatenation of audio and visual features (`A+V'). Each feature combination individually leads to a drop in performance, particularly in AUC, compared to the proposed model that utilizes all features. This suggests that neither cross-modal nor uni-modal information alone is sufficient. Notably, the smallest performance drop occurs when using only video features (`V'), indicating the model’s strong reliance on the video modality. This may explain the performance drop in the RVFA category (Table \ref{tab:cross-manipulations}) compared to FVFA categories.

\subsubsection{Impact of Contrastive Loss}

We also trained our classification model without intra-modal and cross-modal contrastive losses and report the results in Table \ref{tab:feature_set}. The performance improves when contrastive loss is used, indicating its effectiveness in guiding the next-frame prediction module. By pulling together features of real frames and pushing apart those of deepfake frames, contrastive loss helps capture relevant intra-modal and cross-modal features for classification.



\subsection{Impact of Masked-Prediction-based Feature Extraction}

\begin{figure*}[htbp]
    \centering
    \subfloat[Real Visual - Real Audio]{%
        \includegraphics[width=0.48\textwidth]{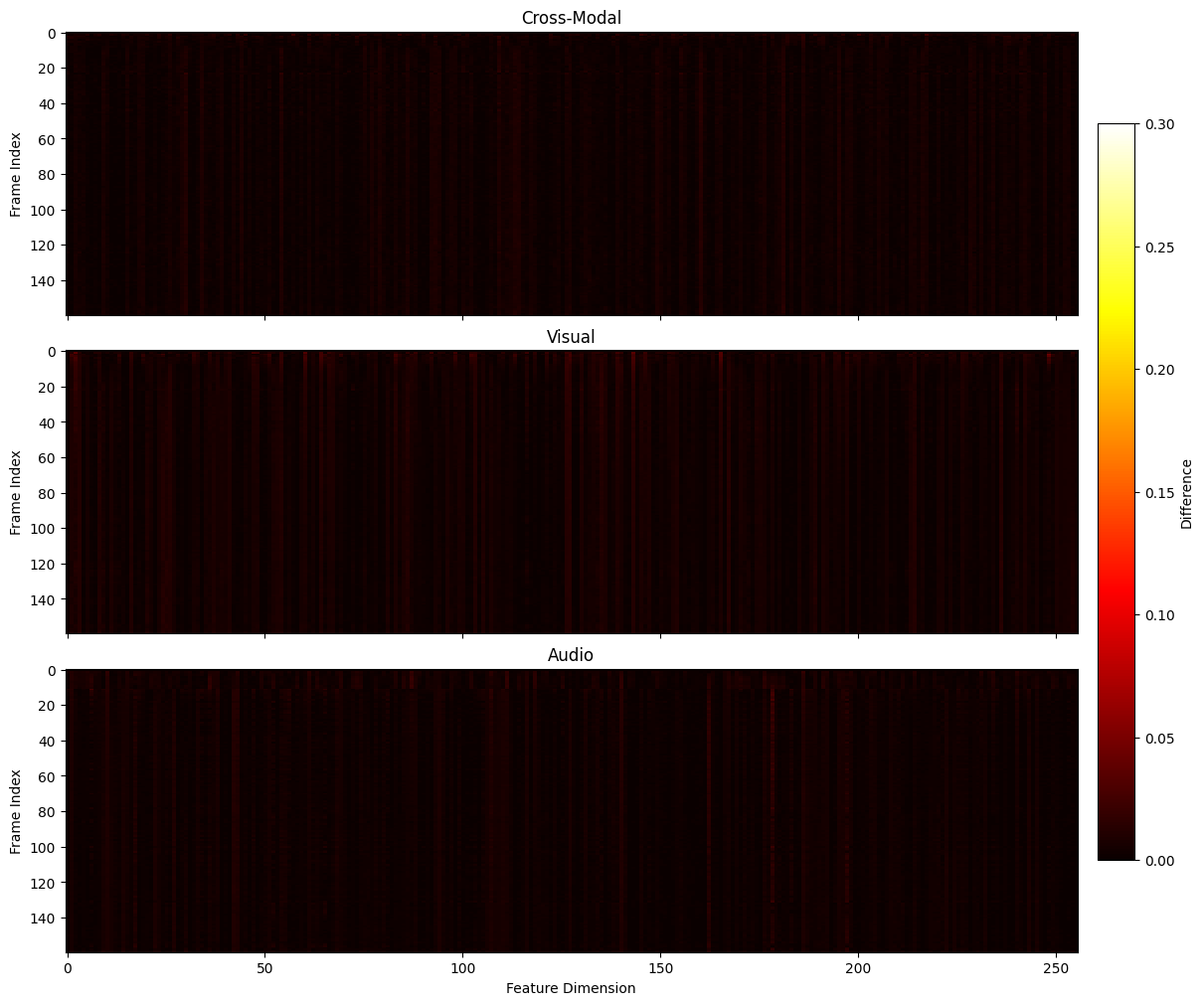}
        \label{fig:hm_RVRA}
    }
    \hfill
    \subfloat[Real Visual - Fake Audio]{%
        \includegraphics[width=0.48\textwidth]{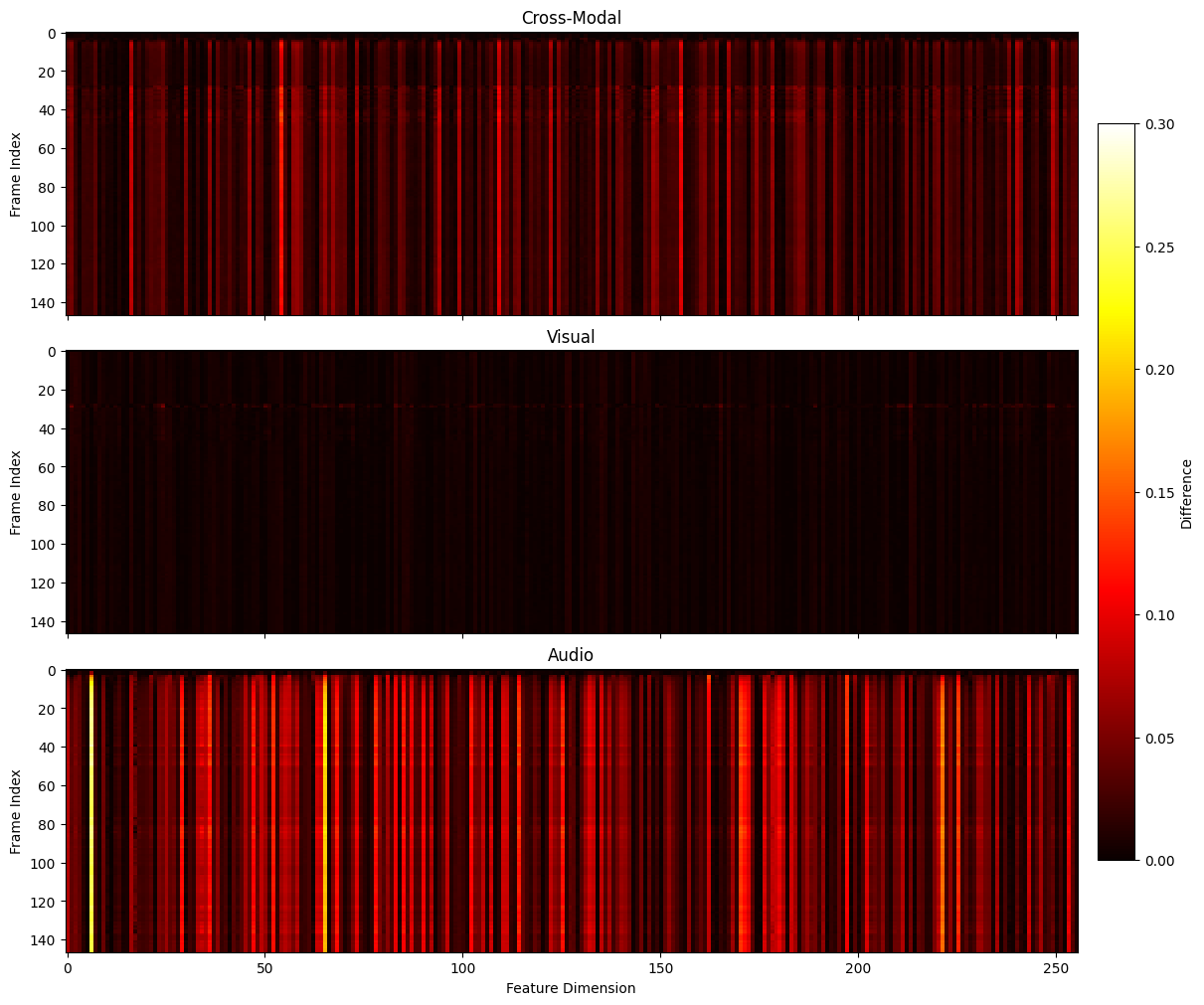}
        \label{fig:hm_RVFA}
    }

    \subfloat[Fake Visual - Fake Audio]{%
        \includegraphics[width=0.48\textwidth]{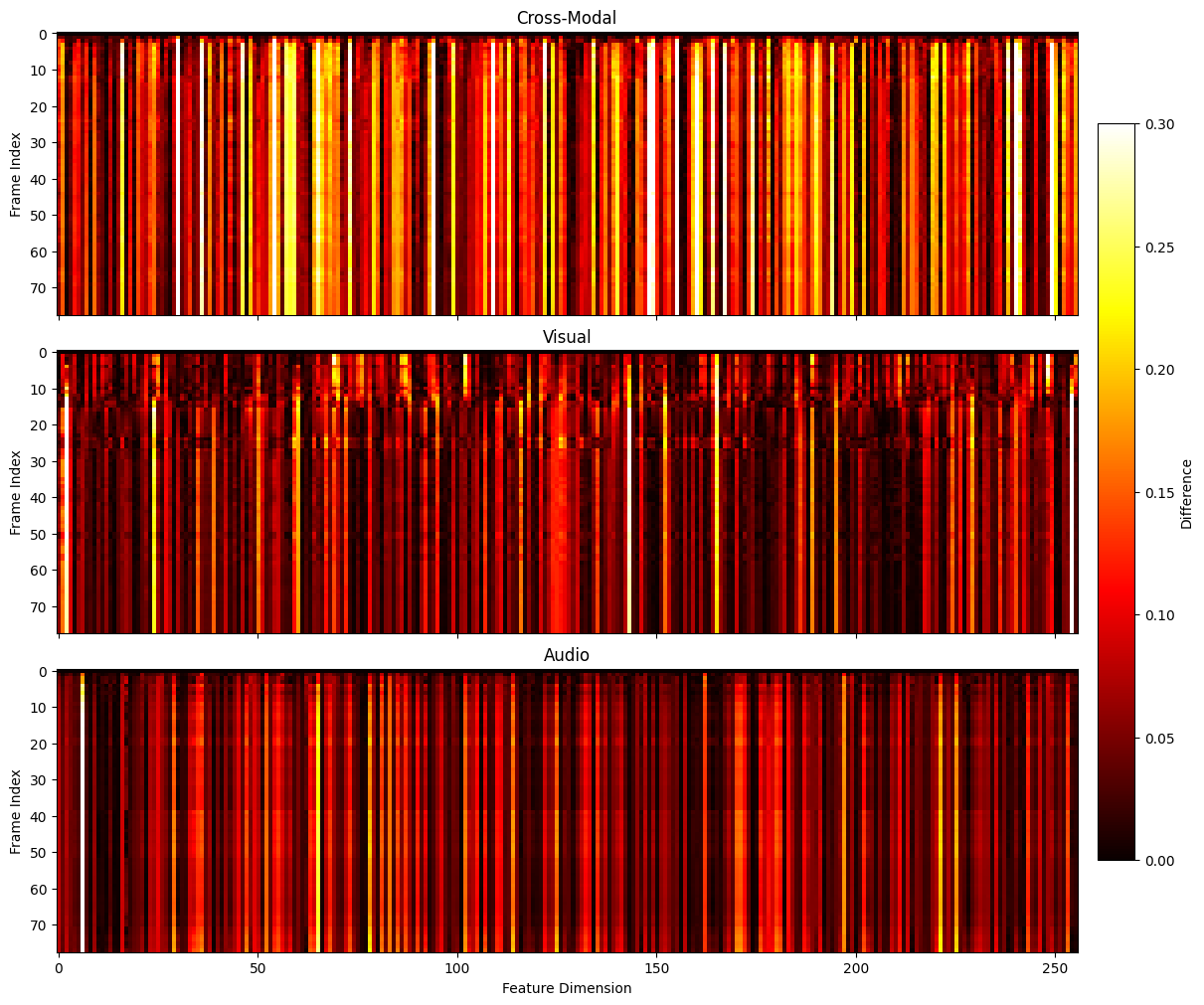}
        \label{fig:hm_FVFA}
    }
    \hfill
    \subfloat[Partial Deepfake]{%
        \includegraphics[width=0.48\textwidth]{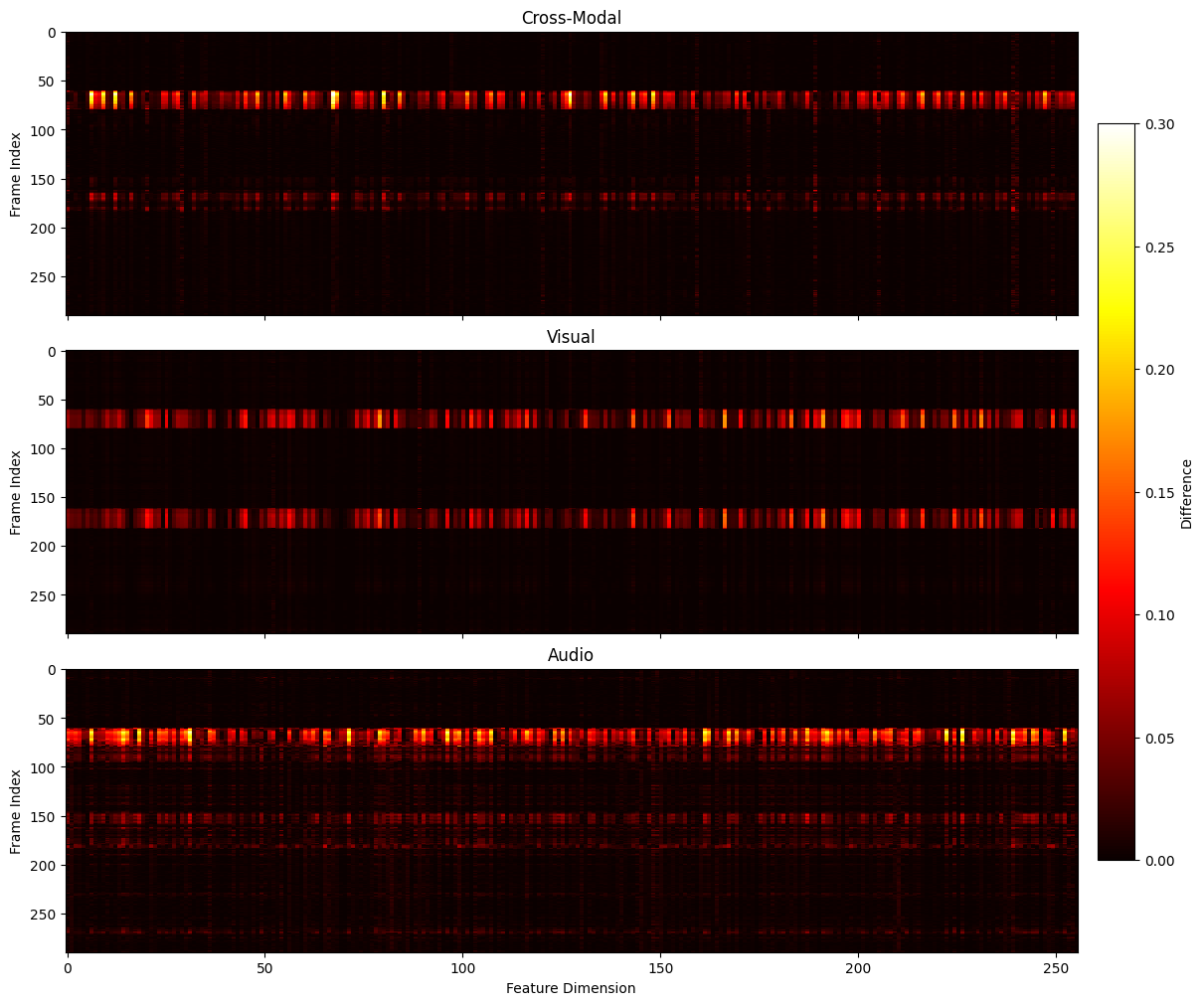}
        \label{fig:hm_lavdf}
    }

    \caption{We process one video sample from each category: (a) Real Visual–Real Audio, (b) Real Visual–Fake Audio, (c) Fake Visual–Fake Audio, and (d) Partial Deepfake through the trained models. We extract intra- and cross-modal frame features, compute absolute pointwise differences, and visualize them as heatmaps for analysis.}
    \label{fig:all_heatmaps}
\end{figure*}

The Masked-Prediction-based Feature Extraction module captures the difference between predicted and actual intra-modal and cross-modal features using convolutional attention, and then passes these differences for classification or temporal localization. We analyze these differences to understand how the model processes real videos and various types of deepfakes. We use four video samples: (i) Real Visual–Real Audio (no manipulation across the video), (ii) Real Visual–Fake Audio (one modality manipulated across the video), (iii) Fake Visual–Fake Audio (both modalities manipulated across the video), and (iv) Partial Deepfake (both modalities manipulated but only in small, non-continuous segments). These are passed through the trained classification model (for samples i–iii) and the localization model (for sample iv). We then extract the actual and predicted intra-modal and cross-modal frame features, which are outputs of the causal transformer encoder and decoder blocks, respectively. Finally, we compute the absolute pointwise difference between three feature pairs (cross-modal, visual, and audio) and plot a heatmap of these differences for further analysis.

Figure~\ref{fig:all_heatmaps} shows the heatmaps for the four samples. Each sample has three heatmaps corresponding to Cross-modal, Visual, and Audio features, ordered from top to bottom. To simplify comparison, the difference scale is fixed across all 12 heatmaps, ranging from 0.0 to 0.3. As expected, the Real Visual–Real Audio sample shows minimal difference between predicted and actual features across all modalities (Figure~\ref{fig:hm_RVRA}). The Real Visual–Fake Audio sample shows minimal difference in the visual modality but clear differences in the audio and cross-modalities (Figure~\ref{fig:hm_RVFA}). The Fake Visual–Fake Audio sample shows visible differences across all modalities (Figure~\ref{fig:hm_FVFA}). In contrast, the Partial Deepfake sample shows two thin stripes of differences in specific temporal ranges across all modalities (Figure~\ref{fig:hm_lavdf}). These heatmaps indicate that the differences reveal which modality, and which temporal ranges, contain manipulations, even when manipulations occur in multiple segments.

The advantages of this analysis are threefold: 
\begin{enumerate}
    \item it demonstrates the positive contribution of the proposed masked prediction-based feature extraction approach to deepfake detection and temporal localization;
    \item it improves model interpretability, as the differences highlight which modality or frame ranges are modified; and
    \item it supports the contribution of combining cross-modal and intra-modal features. As seen in Figure~\ref{fig:hm_RVFA}, where only audio is manipulated, the heatmaps show minimal visual difference, a striking audio difference, and a mild cross-modal difference. Similarly, in Figure~\ref{fig:hm_lavdf}, both stripes are clearly visible in the visual modality, while the bottom stripe appears only mildly in the cross-modal and audio modalities. This shows that when cross-modal differences are less visible and could be missed by the detector, intra-modal differences can aid in detecting manipulations.
\end{enumerate}

\begin{figure}
    \centering
    \includegraphics[width=\linewidth]{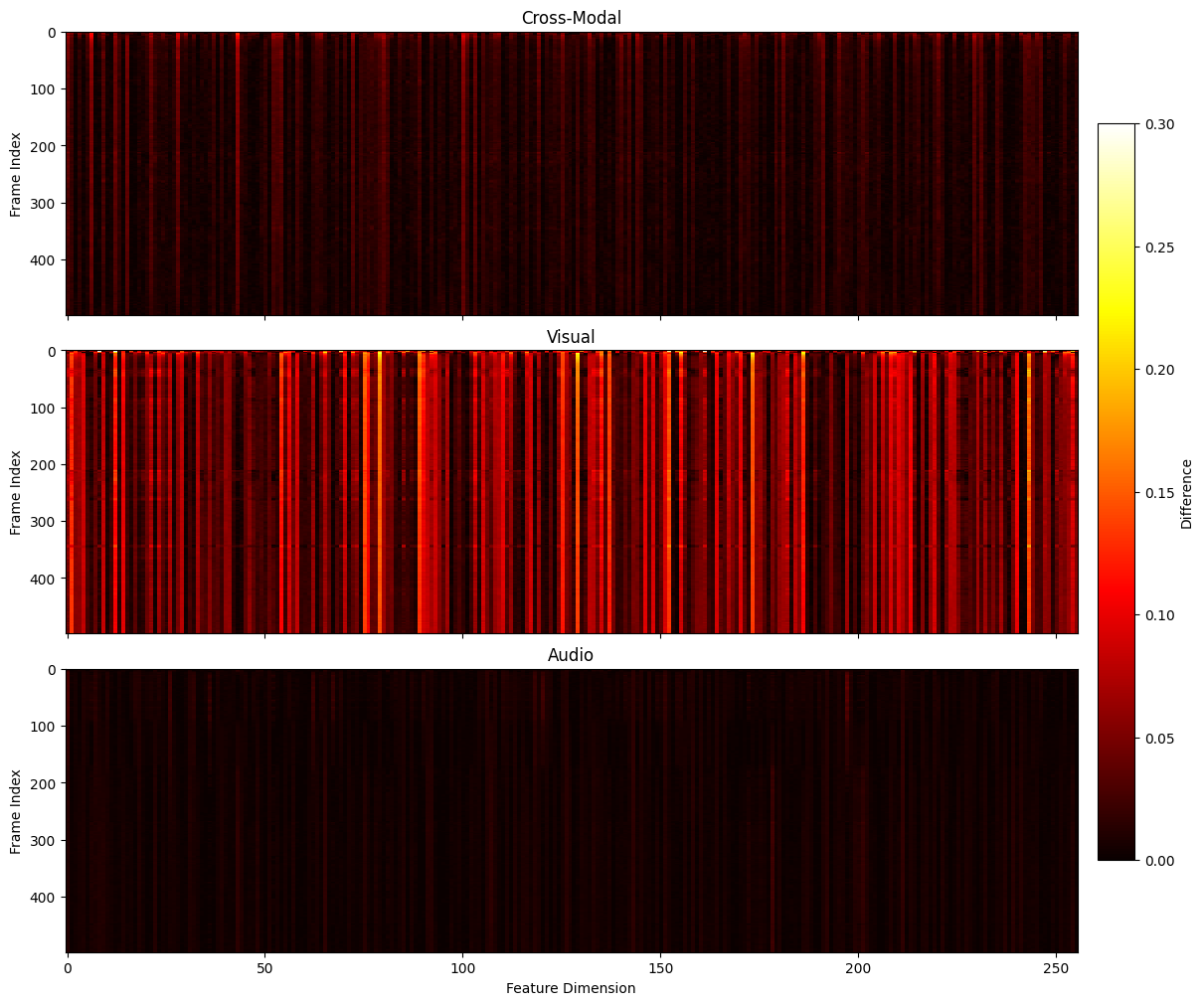}
    \caption{We analyse the heatmaps generated by a sample from the KoDF dataset. They show a significant difference between the predicted and actual features for visual modality, no difference for the audio modality, and a mild difference for the cross-modal features.}
    \label{fig:kodf}
\end{figure}

Additionally, to further investigate the perfect cross-dataset AP and AUC scores achieved on the KoDF dataset, we analyzed heatmaps generated by the model trained on FakeAVCeleb over a KoDF deepfake sample belonging to the Fake Visual–Real Audio category. As shown in Figure~\ref{fig:kodf}, the model correctly predicts higher differences in the visual modality, mild differences in the cross-modality, and minimal differences in the audio modality. This demonstrates that the model’s ability to capture differences between predicted and actual intra-modal and cross-modal features generalizes to unseen datasets, leading to strong performance on KoDF.

\subsection{Complexity Analysis}

\begin{table}
    \centering
    \caption{\textbf{Computational analysis:} We report the inference time and computational complexity in terms of GFLOPs for individual tasks of the proposed model.}
    \label{tab:compute_complex}
    \begin{tabular}{r c c c}
        \toprule
        Tasks & Train Params (M) & GFLOPs & Time (ms) \\
        \midrule
        Classification & 107.61 & 179.44 & 29.87 \\
        Temporal Localization & 127.23 & 209.92 & 69.04 \\
        \bottomrule
    \end{tabular}
\end{table}

To assess the computational complexity of our model, we separately measured the inference time and FLOPs for the classification and localization tasks, as presented in Table~\ref{tab:compute_complex}. All measurements were taken using a batch size of 1 on an RTX A6000 Ada 48GB GPU. The classification module is highly efficient, requiring minimal computation time. Lastly, the reported inference time for localization includes the overhead from non-maximum suppression.

\section{Conclusion}
\label{sec:conclusion}

We propose a single-stage framework for audio-visual deepfake detection and temporal localization, using a shared backbone for both tasks. Our approach leverages masked next-frame feature prediction and convolution-based attention between predicted and actual frame features to capture intra- and cross-modal inconsistencies. Despite its single-stage design, our model demonstrates strong cross-manipulation and cross-dataset performance. Additionally, the same backbone effectively supports audio-visual deepfake temporal localization. A promising future direction is the joint training of detection and localization.

\bibliographystyle{IEEEtran}
\bibliography{main}

\end{document}